\documentclass{article}
\usepackage[margin=1in]{geometry}
\usepackage{adjustbox}
\usepackage{amssymb,amsmath}
\usepackage{graphicx}
\usepackage{cite}
\usepackage{hyperref}
\usepackage[numbers,sort&compress]{natbib}
\usepackage{subcaption}
\usepackage{placeins}
\usepackage{longtable}
\usepackage{booktabs}   

\hypersetup{
    colorlinks=true,
    linkcolor=blue,
    citecolor=blue,
    urlcolor=blue
}
\newcommand{\ie}{{\sl i.e.}~}

\title{Toward Ethical AI Through Bayesian Uncertainty in Neural Question Answering}
\author{Riccardo Di Sipio \\ \small Dayforce, HCM \\ \small \texttt{riccardo.disipio@dayforce.com}}
\date{\today}

\begin{document}

\maketitle

\begin{abstract}
We explore Bayesian reasoning as a means to quantify uncertainty in neural networks for question answering. Starting with a multilayer perceptron on the Iris dataset, we show how posterior inference conveys confidence in predictions. We then extend this to language models, applying Bayesian inference first to a frozen head and finally to LoRA-adapted transformers, evaluated on the CommonsenseQA benchmark. Rather than aiming for state-of-the-art accuracy, we compare Laplace approximations against maximum a posteriori (MAP) estimates to highlight uncertainty calibration and selective prediction. This allows models to abstain when confidence is low. An "I don't know" response not only improves interpretability but also illustrates how Bayesian methods can contribute to more responsible and ethical deployment of neural question-answering systems. 
\end{abstract} 

\section{Introduction}
\label{sec:intro}

Large neural networks have achieved remarkable progress in natural language processing, particularly in tasks such as question answering \cite{devlin2019bert,liu2019roberta,raffel2020t5,brown2020gpt3}. Yet despite their predictive power, these models typically produce point estimates without a measure of confidence \cite{guo2017calibration,desai2020calibration,gal2016dropout,kendall2017uncertainties}. This absence of calibrated uncertainty can be problematic: models may answer confidently even when wrong, which is particularly concerning in high-stakes applications \cite{kamath2020selectivity,jiang2021knows,amodei2016concrete,oneill2016weapons}.

\paragraph{}
In a standard neural network, training produces a single set of parameters~$\theta$, and predictions are made as point estimates 
$p(y|x,\theta)$. In contrast, Bayesian reasoning treats the parameters themselves as random variables with a distribution that captures 
our uncertainty. A prior distribution $p(\theta)$ encodes initial beliefs, the likelihood $p(y|x,\theta)$ links data to parameters, 
and Bayes’ rule gives a posterior $p(\theta|{\cal D}) \propto p({\cal D}|\theta)p(\theta)$ after observing data ${\cal D}$. 
Predictions then marginalize over this posterior:
\[
p(y|x,{\cal D}) = \int p(y|x,\theta)\, p(\theta|{\cal D}) \, d\theta.
\]
This marginalization yields predictive distributions that reflect both the data fit and the epistemic uncertainty of the model. 
In practice, exact posteriors are intractable for large networks, so approximations such as Laplace \cite{mackay1992laplace}~or Monte Carlo \cite{neal2012bayesian,robert2004monte}~sampling 
are employed. 

\paragraph{}
Thus, Bayesian reasoning provides a principled framework for addressing the lack of calibrated uncertainties in neural networks \cite{blundell2015weight}. By combining priors with likelihoods to form posteriors, Bayesian methods not only make predictions but also quantify the level of belief in those predictions. This opens the possibility for models to abstain \ie to say “I don’t know” when confidence is low. Such behavior is not only useful for downstream tasks like selective prediction and calibration, but also central to building systems that align with the goals of responsible and ethical AI \cite{floridi2019unified,jobin2019global,mitchell2019modelcards,oneill2016weapons}.
\paragraph{}
In this paper, we explore this idea through three experiments of increasing complexity:

\begin{enumerate}
    \item A simple baseline (Iris dataset): We revisit a multilayer perceptron trained with Bayesian inference via MCMC sampling, showing how posterior predictive distributions provide a natural notion of uncertainty.
    \item Bayesian head on a frozen language model: We apply Bayesian inference to the classification head of a pretrained transformer while keeping the backbone frozen.
    \item Finally, we extend the approach by fine-tuning transformer adapters with LoRA \cite{hu2022lora}~and applying a Laplace approximation over adapter and head parameters. We evaluate this setup on the CommonsenseQA \cite{talmor2019commonsenseqa}~benchmark, focusing not on state-of-the-art accuracy but on the quality of uncertainty estimates.
\end{enumerate}

Across these experiments, we illustrate how Bayesian posteriors can inform reliability diagrams, selective prediction, and per-example uncertainty analysis. While the emphasis is educational rather than competitive, the results underscore how Bayesian treatments can enrich neural question answering with calibrated uncertainty and abstention, laying groundwork for broader applications in generative AI.

\section{Related work}
\label{sec:related_work}

{\bf Question Answering Benchmarks.} Commonsense reasoning has become a standard testbed for evaluating language models beyond surface-level pattern matching. The CommonsenseQA dataset introduced by Talmor et al.\cite{talmor2019commonsenseqa}~provides multiple-choice questions designed to probe background knowledge and reasoning ability.

{\bf Parameter-Efficient Fine-Tuning.} Transformer-based models such as BERT \cite{devlin2019bert} have motivated methods for efficient adaptation to downstream tasks. LoRA (Low-Rank Adaptation) introduced by Hu et al.\cite{hu2022lora}~injects trainable low-rank matrices into frozen weights, reducing memory and compute while retaining performance. LoRA has since been applied widely in large language models, including for uncertainty-aware training \cite{he2022towards,karampatziakis2023uncertainty}.

{\bf Bayesian Neural Networks.} Bayesian inference for neural networks has a long history, including exact posterior sampling via Markov Chain Monte Carlo (MCMC) \cite{neal2012bayesian}, variational inference, and Laplace approximations \cite{mackay1992laplace}. These approaches allow uncertainty quantification by treating weights as random variables rather than fixed parameters. Recent work has revisited Laplace approximations in modern deep learning contexts, demonstrating their effectiveness for calibration and predictive uncertainty \cite{daxberger2021laplace}.

{\bf Uncertainty in NLP.} A growing body of work explores Bayesian and calibration methods in natural language processing, including applications to classification and question answering \cite{xiao2020bayesianbert,malinin2020uncertaintyqa,gal2016dropout}. Reliability diagrams, selective prediction, and abstention mechanisms have been studied as tools to align model confidence with empirical accuracy \cite{guo2017calibration}. Our work contributes to this line by adapting Bayesian methods to transformer-based QA, with a particular emphasis on interpretability and abstention.

\section{Experimental Demonstrations}
\label{sec:experiments}
To make the discussion concrete, we present three experimental demonstrations of increasing complexity. Each experiment illustrates how Bayesian inference enriches neural networks with calibrated uncertainty, moving from a pedagogical baseline to modern language models for question answering. The emphasis is educational rather than competitive, with a focus on interpretability, abstention, and ethical deployment rather than state-of-the-art accuracy.

\subsection{Experiment 1: Bayesian Inference on the Iris Dataset}
\label{sec:iris}

As a pedagogical starting point, we revisit Bayesian inference on the classic Iris dataset \cite{fisher1936iris}, demonstrating how posterior predictive distributions reflect uncertainty in classification.

This small, well-structured dataset remains a useful teaching example, consisting of 150 labeled samples across three flower species, each described by four continuous features.

We train a simple multilayer perceptron (MLP) classifier on this dataset, but unlike standard training where weights are optimized to point estimates, we treat the weights as random variables with prior distributions. Using Markov Chain Monte Carlo (MCMC) sampling \cite{neal2012bayesian,robert2004monte}, we draw from the posterior distribution of the weights conditioned on the data.

This Bayesian treatment allows us to:
\begin{itemize}
    \item Visualize priors vs posteriors: showing how the data updates our initial beliefs about parameters.
    \item Obtain predictive distributions: instead of producing a single probability vector, the network integrates over weight samples to generate a distribution over predictions.
    \item Quantify uncertainty: for each test example, we can report not only the predicted class but also the posterior variance, highlighting cases where the model is unsure.
\end{itemize}	

To make these ideas concrete, we reproduce three kinds of results:
\begin{enumerate}
    \item Prior vs posterior plots for selected weights, showing how uncertainty narrows after conditioning on data.
    \item Two-dimensional marginal posterior distributions for weight pairs, which sometimes exhibit correlation or multimodality.
    \item Per-example posterior predictive distributions, where mean predictions are accompanied by error bars. In particular, these plots illustrate cases where one class is most probable but another remains a close runner-up, motivating the idea of an “I don’t know” response when confidence is too low.
\end{enumerate}
	
Rather than aiming for performance improvements on this simple dataset, our goal here is educational: to demonstrate how Bayesian reasoning naturally introduces the concept of belief and uncertainty, setting the stage for more complex models in subsequent experiments.

\paragraph{}
To build intuition for the Bayesian treatment of neural networks, we begin with controlled toy settings where the posterior can be visualized directly. These demonstrations illustrate how priors, likelihoods, and posteriors interact to shape the model’s predictions and associated uncertainty.

\begin{figure*}[htbp]
  \centering
  \begin{subfigure}[t]{0.48\textwidth}
    \centering
    \includegraphics[width=\linewidth]{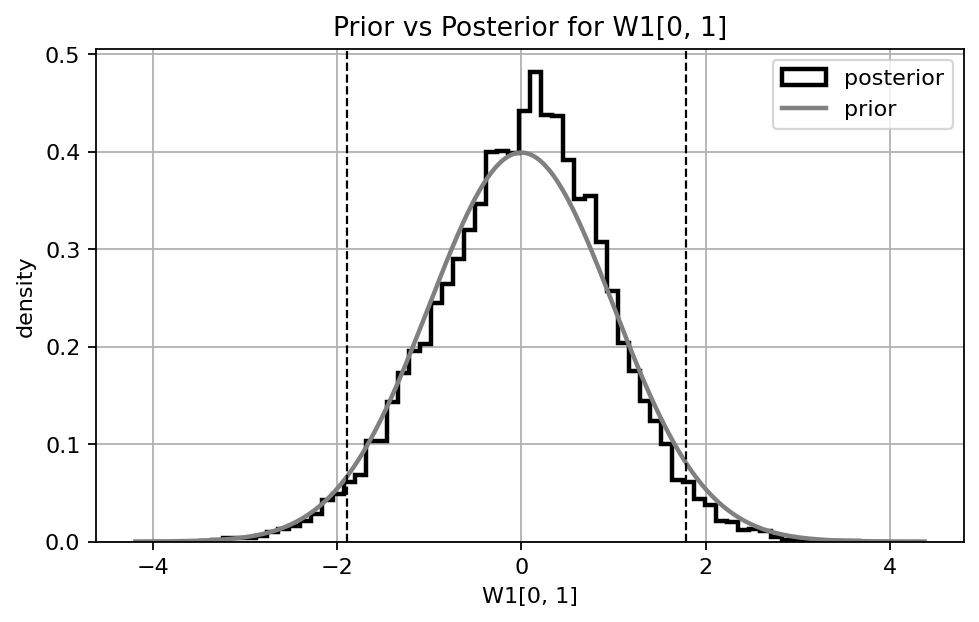}
    \caption{Prior vs.\ posterior for \(W_1[0,1]\).}
  \end{subfigure}\hfill
  \begin{subfigure}[t]{0.48\textwidth}
    \centering
    \includegraphics[width=\linewidth]{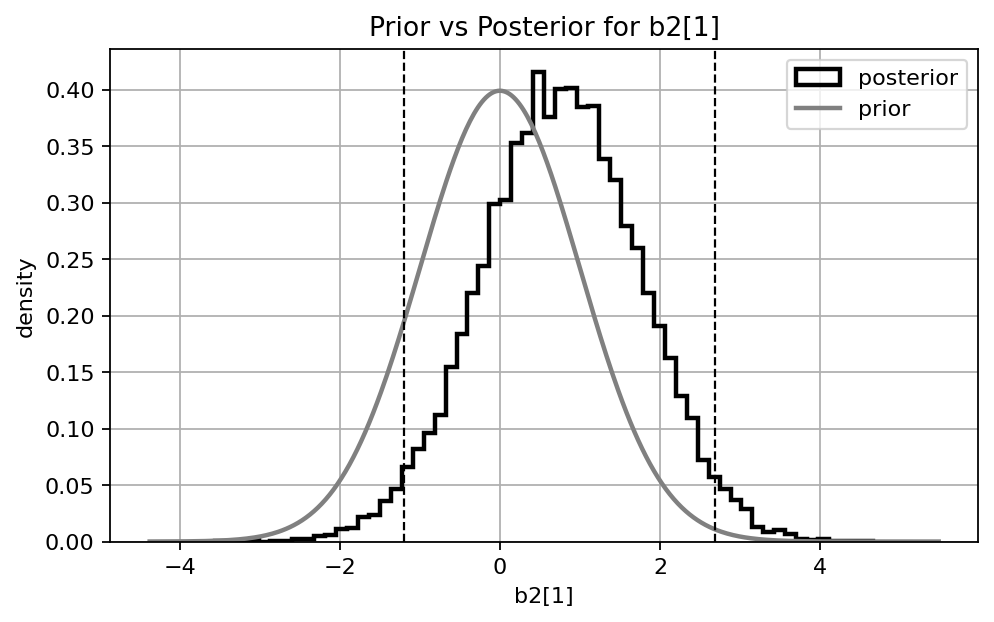}
    \caption{Prior vs.\ posterior for \(b_2[1]\).}
  \end{subfigure}
  \caption{\small{One–dimensional priors (gray) and marginalized posteriors (black) for selected parameters.
  Posteriors concentrate and shift relative to priors as the data updates beliefs.}}
  \label{fig:exp1:1dpost}
\end{figure*}

\paragraph{}
At the one–dimensional level, we can compare prior and posterior distributions over individual parameters of a small network. As shown in Fig.~\ref{fig:exp1:1dpost}, the prior is broad and uninformative, while the posterior becomes more concentrated around values supported by the data. This highlights how Bayesian updating reduces uncertainty when evidence accumulates.

\begin{figure*}[htbp]
  \centering
  \begin{subfigure}[t]{0.48\textwidth}
    \centering
    \includegraphics[width=\linewidth]{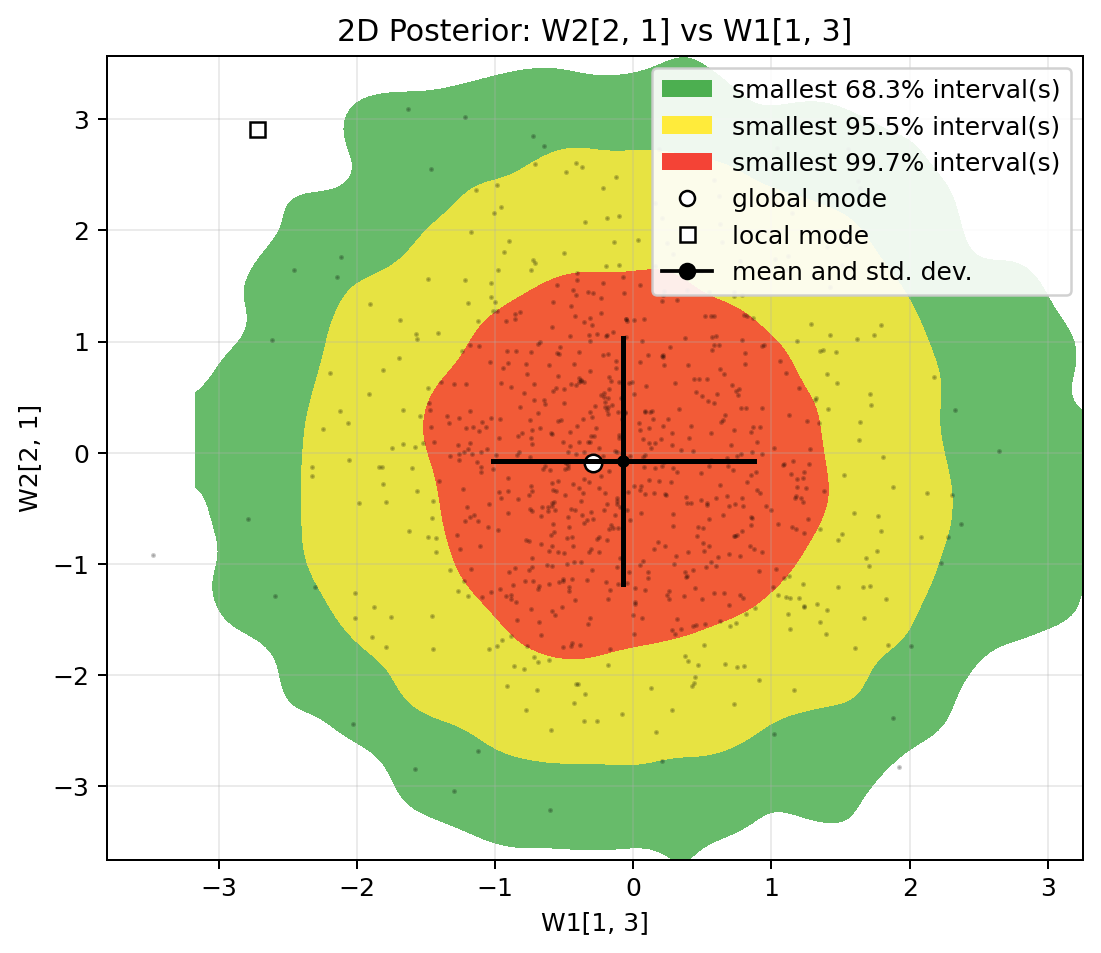}
    \caption{\small{\(W_1[1,3]\) vs.\ \(W_2[2,1]\): near-circular (weak correlation).}}
  \end{subfigure}\hfill
  \begin{subfigure}[t]{0.48\textwidth}
    \centering
    \includegraphics[width=\linewidth]{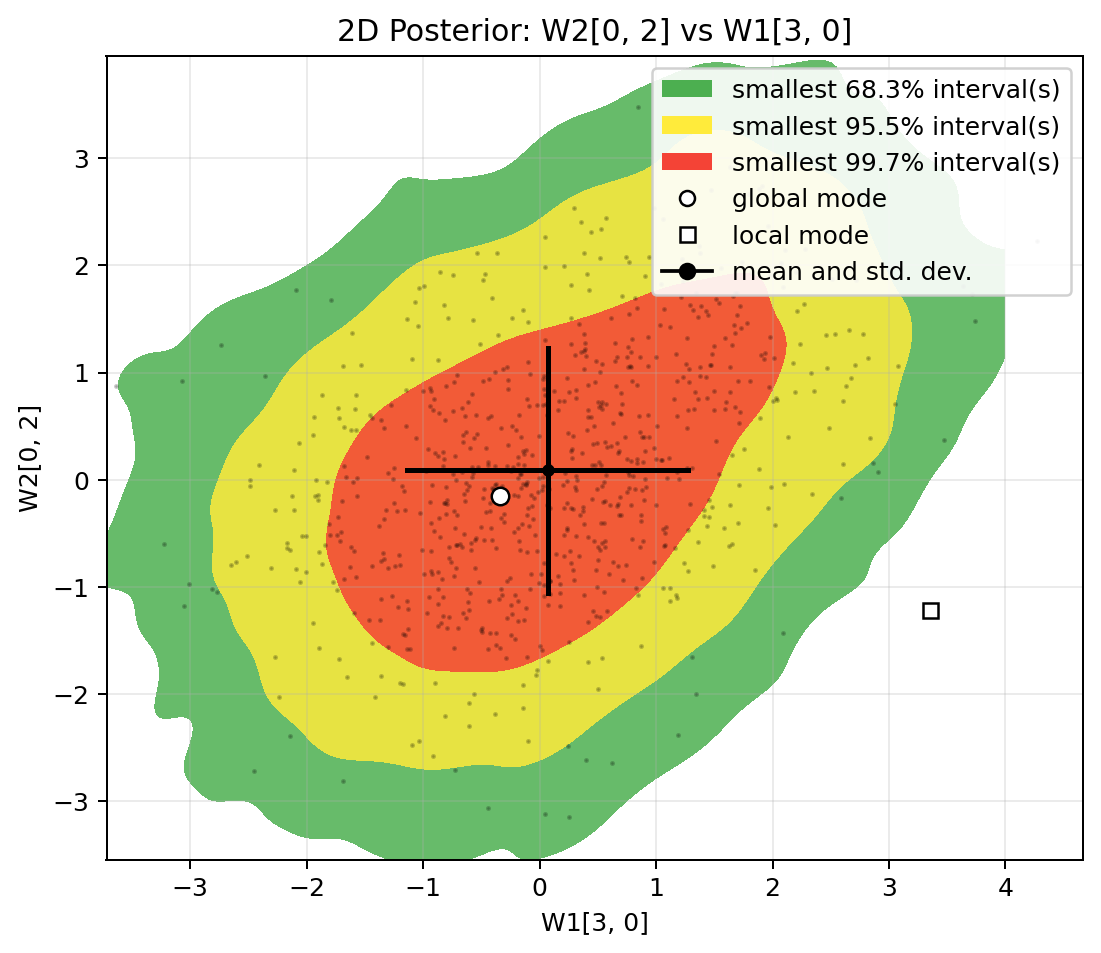}
    \caption{\small{\(W_1[3,0]\) vs.\ \(W_2[0,2]\): tilted contours (correlated).}}
  \end{subfigure}
  \caption{\small{Two–dimensional marginalized posteriors with credible-region contours (68/95/99.7\%).
  Geometry reveals how uncertainty couples parameters.}}
  \label{fig:exp1:2dpost}
\end{figure*}

\paragraph{}
In higher dimensions, posteriors capture not only marginal variance but also correlations between parameters. Fig.~\ref{fig:exp1:2dpost} contrasts two examples: one where parameters remain nearly independent, and another where strong posterior correlation emerges. This illustrates how the geometry of the parameter space is reshaped by data, an effect that is invisible in maximum–likelihood or MAP estimates.

\begin{figure*}[htbp]
  \centering
  \begin{subfigure}[t]{0.48\textwidth}
    \centering
    \includegraphics[width=\linewidth]{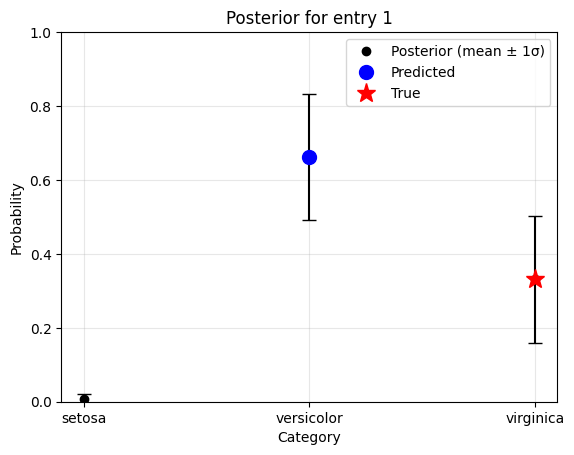}
    \caption{\small{Entry \#1: confident correct prediction.}}
  \end{subfigure}\hfill
  \begin{subfigure}[t]{0.48\textwidth}
    \centering
    \includegraphics[width=\linewidth]{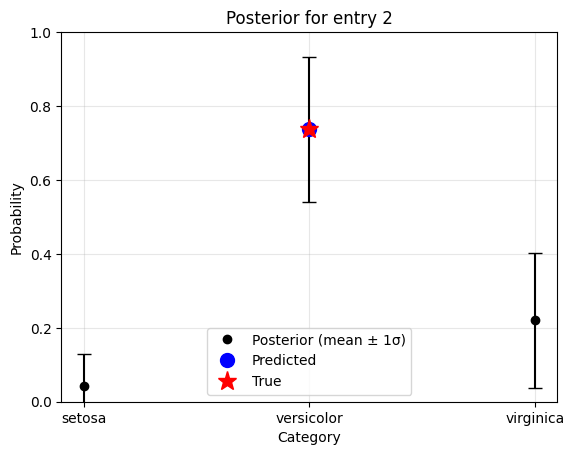}
    \caption{\small{Entry \#2: higher uncertainty (wider error bars).}}
  \end{subfigure}
  \caption{\small{Posterior predictive per sample (mean \(\pm\) 1\(\sigma\)).
  Black dots = means; blue circle = predicted class; red star = true class.}}
  \label{fig:exp1:single}
\end{figure*}

\paragraph{}
The effect of this parameter uncertainty propagates naturally to predictions. Fig.~\ref{fig:exp1:single} shows per–question posterior predictive distributions for individual examples, including the mean probability, one–sigma error bars, the predicted class, and the true label. These plots make explicit when the model is confidently correct, confidently wrong, or uncertain.

\begin{figure*}[htbp]
  \centering
  \begin{subfigure}[t]{0.48\textwidth}
    \centering
    \includegraphics[width=\linewidth]{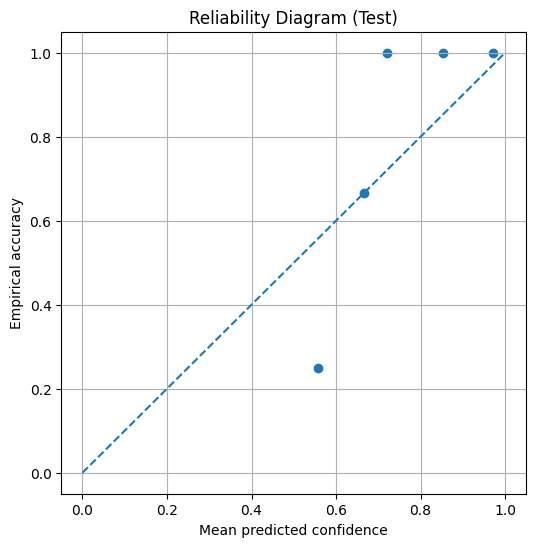}
    \caption{\small{Reliability diagram (test).}}
  \end{subfigure}\hfill
  \begin{subfigure}[t]{0.48\textwidth}
    \centering
    \includegraphics[width=\linewidth]{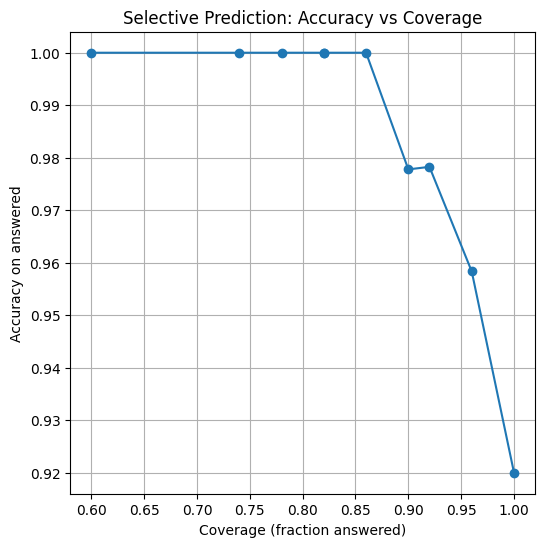}
    \caption{\small{Selective prediction: accuracy vs.\ coverage.}}
  \end{subfigure}
  \caption{\small{System-level evaluation. Left: calibration (confidence vs.\ empirical accuracy).
  Right: accuracy when abstaining below a confidence threshold.}}
  \label{fig:exp1:calib}
\end{figure*}

\paragraph{}
Finally, we examine how posterior predictive uncertainty translates into better calibration and more cautious decision making. Fig.~\ref{fig:exp1:calib} presents two complementary diagnostics. A reliability diagram compares predicted confidence against empirical accuracy: a perfectly calibrated model would fall along the diagonal. An accuracy–coverage curve evaluates selective prediction by plotting the accuracy obtained on answered cases as a function of coverage (the fraction of questions the model chooses to answer when abstaining below a confidence threshold). Together, these metrics show that the Bayesian treatment reduces overconfidence and enables the model to trade coverage for reliability.

\FloatBarrier

\subsection{Experiment 2: A small-scale question answering task with DistilBERT}
\label{sec:frozen_head}

To bridge from toy problems to real benchmarks, we constructed a small synthetic question answering dataset with three answer options per question (examples listed in Appendix). Questions span general knowledge domains such as geography, history, and basic science, with one correct option and two distractors.

\paragraph{}

We use DistilBERT \cite{sanh2019distilbert} to encode each (question, option) pair, extract the \texttt{[CLS]} embedding, and concatenate the resulting three vectors into a single feature representation. This representation is then passed to a Bayesian multinomial logistic regression head, trained using Hamiltonian Monte Carlo (HMC) with the NUTS sampler \cite{hoffman2014no}~as implemented in NumPyro \cite{phan2019numpyro}.

Because the feature dimension is modest (around $7,000$ parameters with $D=768$ for DistilBERT), full posterior sampling is computationally feasible, allowing us to characterize epistemic uncertainty directly rather than through approximations. This setup highlights how a Bayesian treatment can be layered on top of a pretrained encoder, even when resources are limited.

\begin{figure}[htbp]
    \centering
    \begin{subfigure}{0.48\textwidth}
        \includegraphics[width=\linewidth]{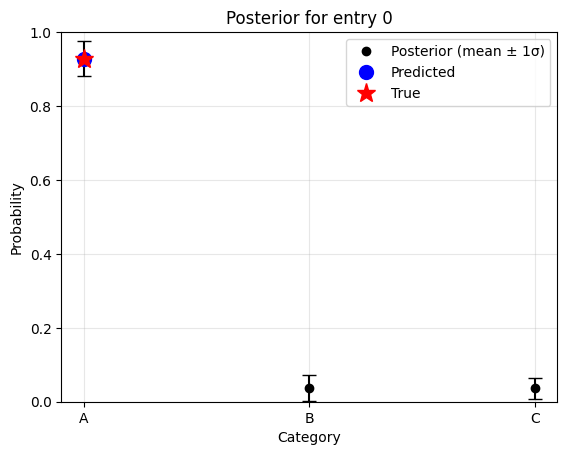}
        \caption{Posterior for entry 0}
    \end{subfigure}
    \hfill
    \begin{subfigure}{0.48\textwidth}
        \includegraphics[width=\linewidth]{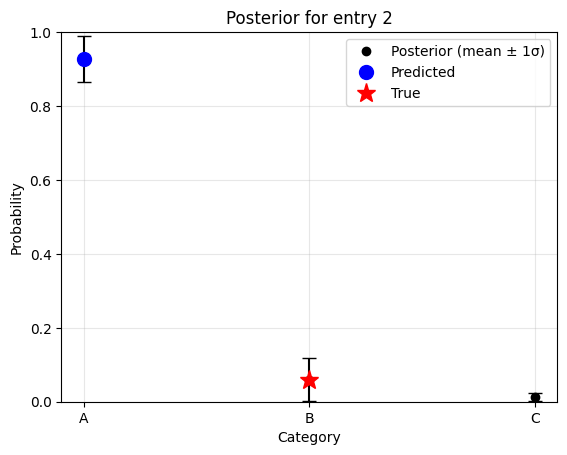}
        \caption{Posterior for entry 2}
    \end{subfigure}

    \begin{subfigure}{0.48\textwidth}
        \includegraphics[width=\linewidth]{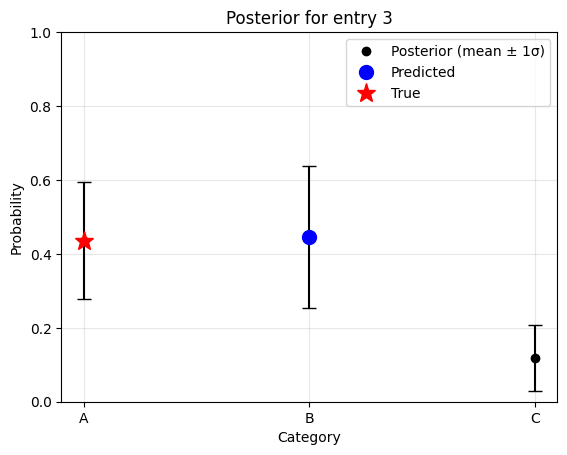}
        \caption{Posterior for entry 3}
    \end{subfigure}
    \caption{\small{Posterior predictive distributions for selected entries in the custom three-class question answering dataset. Black points and error bars show the mean and $\pm 1\sigma$ uncertainty across posterior samples, the blue dot marks the predicted class, and the red star denotes the true label.}}
    \label{fig:exp2:posteriors}
\end{figure}

\paragraph{}
Figure \ref{fig:exp2:posteriors} shows posterior distributions for individual entries in the toy QA dataset. In Entry 0, the model is confident and correct: the predicted answer aligns with the ground truth, and the posterior variance is small. In Entry 3, uncertainty is higher, and the posterior reflects ambiguity between two plausible answers. By contrast, Entry 2 (shown in Fig. \ref{fig:exp2:posteriors}b) illustrates a failure case: the model is highly confident but incorrect. This highlights the importance of evaluating not only accuracy but also calibration and coverage, since confidence alone may mislead users when the model is wrong.

\begin{figure}[htbp]
    \centering
    \begin{subfigure}{0.48\textwidth}
        \includegraphics[width=\linewidth]{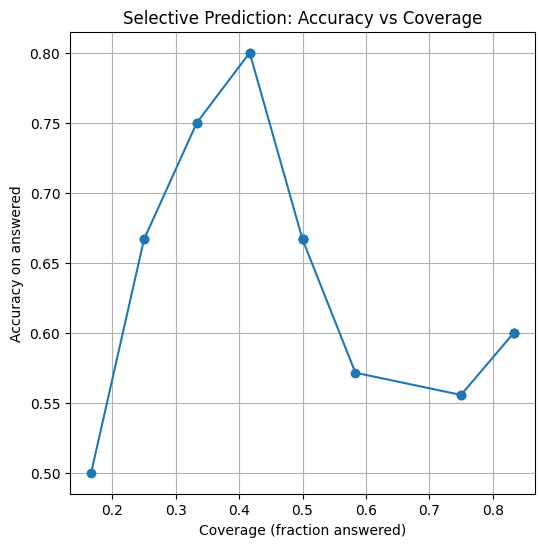}
        \caption{Selective prediction}
    \end{subfigure}
    \hfill
    \begin{subfigure}{0.48\textwidth}
        \includegraphics[width=\linewidth]{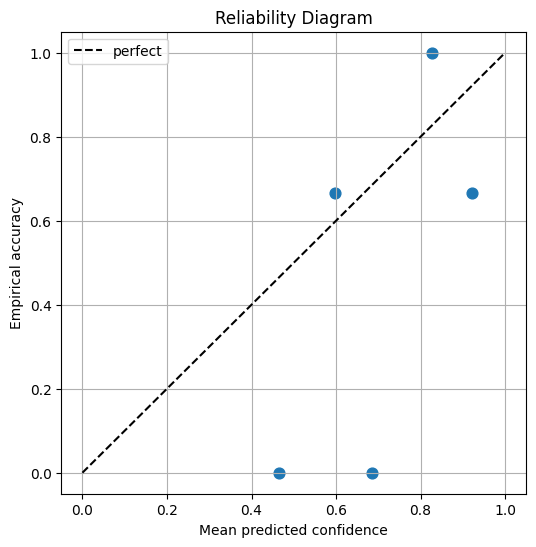}
        \caption{Reliability diagram}
    \end{subfigure}
    \caption{\small{Calibration plots for Experiment~2. (a) Accuracy–coverage trade-off, showing how performance improves when abstaining on low-confidence answers. (b) Reliability diagram, comparing mean predicted confidence with empirical accuracy across bins.}}
    \label{fig:exp2:calibration}
\end{figure}

Beyond single–question posteriors, we also examined aggregate calibration on the toy QA dataset. 
Fig.~\ref{fig:exp2:calibration} (left) shows the accuracy–coverage curve: as the confidence threshold increases, the model abstains on more examples, yielding higher accuracy on the subset it does answer. 
This indicates that the model’s predicted probabilities do carry useful information about uncertainty, even if imperfect. 
The reliability diagram in Fig.~\ref{fig:exp2:calibration} (right) provides a complementary view by comparing predicted confidence against empirical accuracy. 
Although the small dataset and limited training lead to noisy estimates, the plot reveals a tendency toward overconfidence, where predicted probabilities exceed the actual likelihood of correctness. 
These calibration artifacts foreshadow the importance of more principled Bayesian approaches explored in the following section.

\FloatBarrier

\subsection{Experiment 3: LoRA with Laplace Approximation on CommonsenseQA}
\label{sec:laplace_approx}

We scale the Bayesian treatment to a realistic QA benchmark by fine-tuning a pretrained encoder with LoRA adapters and then placing a Bayesian posterior over the small classification head.

We fine-tune {\tt BERT-base-uncased}~\cite{devlin2019bert}~on CommonsenseQA\cite{talmor2019commonsenseqa}~with LoRA adapters applied to the last two transformer layers, targeting the attention projections (query, key, value) and the output dense module. The backbone BERT weights are frozen; only the LoRA parameters and a linear classification head are updated during training. 

\paragraph{}
We approximate the posterior over the head parameters using a \textit{Laplace approximation}. 
Concretely, after training to a maximum a posteriori (MAP) solution, we estimate the local curvature 
of the loss surface via the \textit{empirical Fisher information}. For parameters $\theta$, 
the Fisher is defined as
\[
F(\theta) = \mathbb{E}\!\left[\nabla_\theta \log p(y|x,\theta)\,\nabla_\theta \log p(y|x,\theta)^\top\right],
\]
which captures the sensitivity of the likelihood to changes in $\theta$. In practice, we compute its 
\textit{diagonal approximation} by averaging squared gradients over the dataset. The resulting Gaussian 
posterior, with mean at the MAP parameters and variance given by the inverse Fisher, provides a tractable 
way to sample parameter perturbations and thus quantify predictive uncertainty.

\paragraph{}
In practical terms, posterior predictive distributions are obtained by Monte Carlo sampling, 
typically with $S_{\mathrm{MC}} \equiv 30$, and averaging the resulting softmax outputs. 
This setup balances scalability with the ability to capture epistemic uncertainty in a 
realistic question answering setting.

For efficiency, each question’s five options are reduced to three by sampling two distractors uniformly while preserving the correct answer in a random position. We optionally subsample the training set for faster runs.

\begin{figure}[t]
    \centering
    \begin{subfigure}{0.48\textwidth}
        \includegraphics[width=\linewidth]{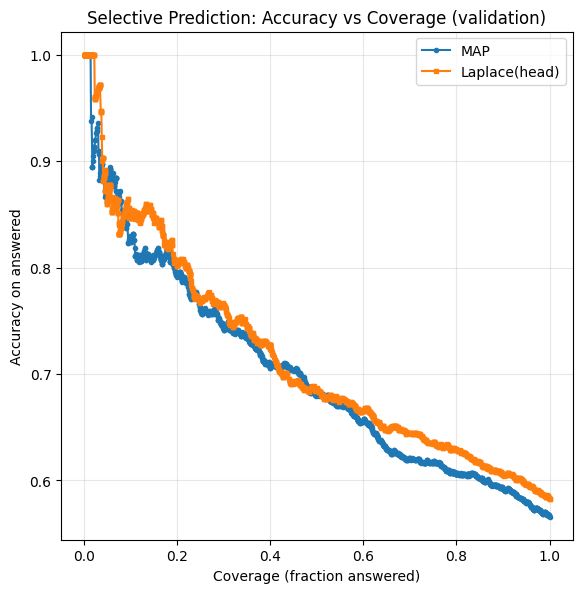}
        \caption{Accuracy vs coverage}
    \end{subfigure}
    \hfill
    \begin{subfigure}{0.48\textwidth}
        \includegraphics[width=\linewidth]{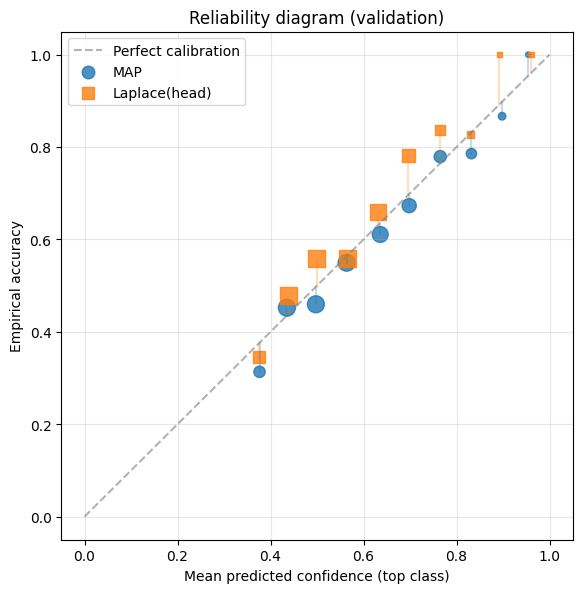}
        \caption{Reliability diagram}
    \end{subfigure}
    \caption{\small Calibration results on CommonsenseQA with \texttt{bert-base-uncased}.
    (a) Accuracy–coverage curves show Laplace slightly outperforming MAP as
    coverage decreases. (b) Reliability diagram compares empirical accuracy against
    predicted confidence, indicating improved calibration under Laplace.}
    \label{fig:exp3:calibration}
\end{figure}

\paragraph{}
The selective prediction curve in Fig.~\ref{fig:exp3:calibration}(a) compares
the accuracy–coverage trade-off for maximum a posteriori (MAP) predictions and
their Laplace-approximated counterparts. As coverage decreases (the model abstains
on low-confidence answers), both curves rise in accuracy, with Laplace consistently
tracking slightly above MAP.

\paragraph{}
Fig.~\ref{fig:exp3:calibration}(b) presents the reliability diagram comparing MAP
and Laplace predictions. Both methods follow the diagonal closely, indicating
reasonable calibration. Laplace does not drastically change the overall shape
of the curve, but it slightly adjusts confidence levels in some bins. The effect
is modest, suggesting that in this setup the primary benefit of Laplace lies
not in large calibration gains but in providing a posterior distribution over
parameters that supports principled uncertainty quantification.

\begin{figure}[t]
    \centering
    \begin{subfigure}{0.32\textwidth}
        \includegraphics[width=\linewidth]{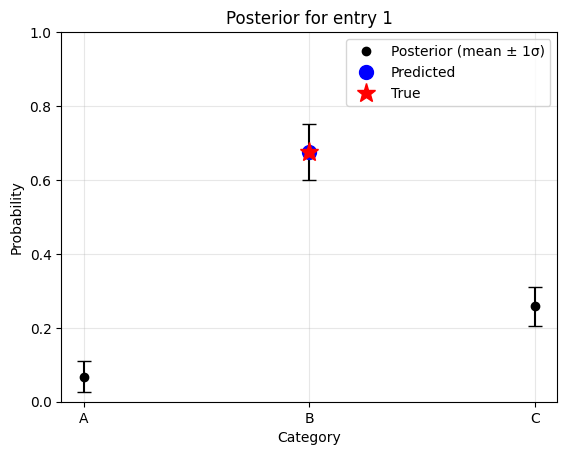}
        \caption{Entry 1}
    \end{subfigure}
    \hfill
    \begin{subfigure}{0.32\textwidth}
        \includegraphics[width=\linewidth]{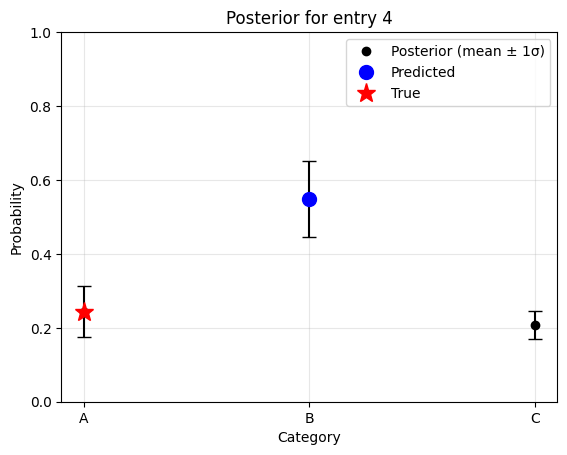}
        \caption{Entry 4}
    \end{subfigure}
    \hfill
    \begin{subfigure}{0.32\textwidth}
        \includegraphics[width=\linewidth]{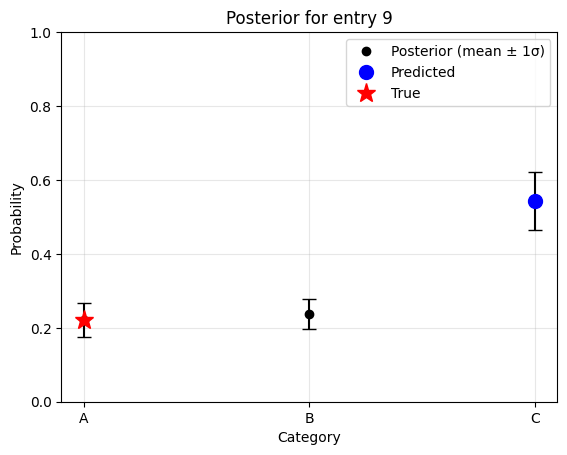}
        \caption{Entry 9}
    \end{subfigure}
    \caption{\small Posterior predictive distributions on CommonsenseQA examples.
    Points indicate mean probabilities with $\pm 1\sigma$ error bars. Blue: predicted
    class; red star: true label.}
    \label{fig:exp3:posteriors}
\end{figure}

\paragraph{}
Finally, Fig.~\ref{fig:exp3:posteriors} illustrates posterior predictive distributions for
three individual test entries. Each plot shows the mean predicted probability
with a $\pm 1\sigma$ interval across Monte Carlo samples. The predicted class is
marked in blue, while the true label is shown in red. These examples highlight
cases where the posterior either reinforces correct predictions with low variance,
or reveals heightened uncertainty when the model is prone to error.

\FloatBarrier

\section{Discussion and Conclusions}
\label{sec:conclusions}

\noindent \textbf{Key Contributions.} In this work we have presented three progressively more complex demonstrations of
Bayesian posteriors applied to neural networks for question answering:
\begin{itemize}
    \item \textbf{Experiment 1:} A didactic example on the Iris dataset, showing how Bayesian posteriors emerge from priors and likelihoods, and how they yield calibrated predictions.
    \item \textbf{Experiment 2:} A small multiple-choice QA dataset with DistilBERT embeddings and a Bayesian logistic regression head, demonstrating that full MCMC inference is feasible for compact parameter sets and produces meaningful uncertainty estimates.
    \item \textbf{Experiment 3:} A realistic benchmark on CommonsenseQA using LoRA-adapted \texttt{bert-base-uncased} with a Laplace approximation over the head, illustrating how Bayesian uncertainty can be integrated into modern QA systems.
\end{itemize}

Across these experiments, we consistently observed that Bayesian methods enrich the
outputs of neural networks by associating predictions with principled measures of
uncertainty. A Laplace-based posterior sampling provides distributions that can be used
for selective prediction, improved calibration, and transparent abstention (“I don’t
know”) in ambiguous cases. Such properties are central to the responsible deployment
of AI systems, especially in question answering where confidently incorrect answers
can have undesirable consequences.

This work does not aim for state-of-the-art performance on CommonsenseQA or other
benchmarks. Instead, our goal has been to highlight how Bayesian reasoning can be
applied in practice, bridging statistical foundations with modern neural architectures.
The results suggest that lightweight Bayesian layers, whether via MCMC on compact heads
or Laplace approximations on adapted transformers, are viable strategies for uncertainty-aware
AI. 

Future work could explore richer priors, scalable approximate inference methods,
and downstream tasks where abstention has clear value, such as education or human-AI
collaboration. More broadly, we view this line of research as part of the effort
to align machine learning systems with ethical principles: a model that can quantify
and communicate its own uncertainty is better positioned to support trustworthy
decision making.

\section*{Funding}  
The author did not receive support from any organization for the submitted work.  

\section*{Ethics Declarations}  
The author declares that no ethical approval was required for this study, and that there are no conflicts of interest.

\section*{Acknowledgments}

This preprint has not undergone peer review or any post-submission improvements or corrections. The Version of Record of this article is published in \textit{AI and Ethics}~(Springer Nature), and is available online at \url{https://link.springer.com/article/10.1007/s43681-025-00838-x}{DOI: 10.1007/s43681-025-00838-x}~.

\paragraph{}
The idea for this work was triggered by an interview with Yoshua Bengio published in \textit{La Repubblica}, but is part of a broader conversation on responsible AI and a broader personal engagement with the themes of uncertainty, trust, and the ethical deployment of machine learning systems. It also builds on the author's past experience in experimental particle physics at CERN, where Bayesian methods were routinely used to quantify uncertainty in high-energy physics experiments.


\appendix

\FloatBarrier

\section*{Appendix}

\label{app:toyqa}

To complement Experiment~2, we created a small synthetic dataset of 30 multiple-choice questions with three candidate answers each. This dataset spans general knowledge domains (geography, history, science) and is designed to test the ability of a lightweight model to capture uncertainty in question answering.  

\begin{small}
\setlength{\tabcolsep}{6pt} 
\renewcommand{\arraystretch}{1.2}
\begin{longtable}{p{6cm}p{2.5cm}p{2.5cm}p{2.5cm}c}
\caption{Toy multiple-choice QA dataset used in Experiment~2. Each question has three options (A–C) with the correct label indicated.}
\label{tab:toyqa} \\
\toprule
\textbf{Question} & \textbf{Option A} & \textbf{Option B} & \textbf{Option C} & \textbf{Label} \\
\midrule
\endfirsthead
\toprule
\textbf{Question} & \textbf{Option A} & \textbf{Option B} & \textbf{Option C} & \textbf{Label} \\
\midrule
\endhead
Which planet is known as the Red Planet? & Mars & Venus & Jupiter & 0 \\
Capital of France? & Paris & Berlin & Madrid & 0 \\
Which animal barks? & dog & cat & cow & 0 \\
Which country hosted the 2016 Summer Olympics? & Brazil & China & UK & 0 \\
Who discovered penicillin? & Alexander Fleming & Marie Curie & Louis Pasteur & 0 \\
What is the capital of Japan? & Kyoto & Tokyo & Osaka & 0 \\
Which is the fastest land animal? & Cheetah & Horse & Lion & 0 \\
Who wrote `Romeo and Juliet'? & William Shakespeare & Charles Dickens & Mark Twain & 0 \\
Which gas is essential for respiration? & Oxygen & Carbon monoxide & Helium & 0 \\
Which continent is Egypt located in? & Africa & Asia & Europe & 0 \\
\midrule
What color are bananas when ripe? & red & yellow & blue & 1 \\
How many continents are there? & Five & Seven & Six & 1 \\
Who painted the Mona Lisa? & Michelangelo & Leonardo da Vinci & Raphael & 1 \\
What is the boiling point of water at sea level (°C)? & 90 & 100 & 110 & 1 \\
2 + 2 equals? & 3 & 4 & 5 & 1 \\
How many players are on a standard soccer team (on field)? & 9 & 11 & 12 & 1 \\
Which element has the symbol `O'? & Osmium & Oxygen & Gold & 1 \\
Which shape has three sides? & Square & Triangle & Pentagon & 1 \\
What is the largest mammal? & Elephant & Blue Whale & Giraffe & 1 \\
Which ocean is the largest? & Pacific Ocean & Atlantic Ocean & Indian Ocean & 1 \\
\midrule
Which organ pumps blood in the human body? & Lungs & Brain & Heart & 2 \\
The Sun is a ... & planet & comet & star & 2 \\
Which metal is liquid at room temperature? & Mercury & Iron & Aluminum & 2 \\
The Great Wall is located in which country? & India & China & Japan & 2 \\
Which planet has the most moons? & Jupiter & Saturn & Neptune & 2 \\
Which gas do humans exhale? & Oxygen & Carbon dioxide & Nitrogen & 2 \\
What is the chemical symbol for gold? & Ag & Au & Pb & 1 \\
Which city is known as the Big Apple? & New York & Los Angeles & Chicago & 0 \\
Which country is both in Europe and Asia? & Turkey & Spain & Mexico & 0 \\
Which month has 28 days? & February & June & November & 0 \\
\bottomrule
\end{longtable}
\end{small}

\clearpage
\FloatBarrier
\bibliographystyle{unsrt}
\bibliography{bibliography}

\end{document}